\documentclass[runningheads]{llncs}
\usepackage[T1]{fontenc}
\usepackage{graphicx}
\usepackage{amsmath}
\usepackage{amssymb}
\usepackage{booktabs}
\usepackage{tabularx}
\usepackage{array}
\newcolumntype{Y}{>{\centering\arraybackslash}X}

\usepackage{comment}

\usepackage{tikz}
\newcommand{\smallangle}[2]{%
  \tikz[baseline=(O.base), scale=0.3]{
    \coordinate (O) at (0,0);
    \coordinate (A) at (1,0);
    \coordinate (B) at ({cos(#1)}, {sin(#1)});
    \draw[line width=0.4pt] (O) -- (A);
    \draw[line width=0.4pt] (O) -- (B);
    \draw[line width=0.4pt] (0.4,0) arc (0:#1:0.4);
    \node at (0.55,0.2) {\scriptsize$#2$};
  }%
}

\begin{document}
\title{Pointing-Guided Target Estimation via Transformer-Based Attention}

%
%


\author{Luca Müller\inst{1}\orcidID{0009-0000-7682-1786}  \and
Hassan Ali\inst{1}\orcidID{0000-0001-9907-1834}
\and
Philipp Allgeuer\inst{1}\orcidID{0000-0002-2355-0764} 
\and
Lukáš Gajdošech\inst{2}\orcidID{0000-0002-8646-2147} 
\and
Stefan Wermter\inst{1}\orcidID{0000-0003-1343-4775}\thanks{This research was supported by Horizon Europe TERAIS (Grant 101079338), the DFG through the Crossmodal Learning (TRR-169), and the EU TRAIL. We thank Matthias Kerzel for his insightful comments that helped improve this manuscript.}}

%
\authorrunning{L. Müller et al.}
%
\institute{University of Hamburg, Department of Informatics, Knowledge Technology Research Group, Hamburg, Germany\\
\email{\{luca.mueller, hassan.ali, philipp.allgeuer, stefan.wermter\}@uni-hamburg.de} \and
Faculty of Mathematics, Physics and Informatics, Comenius University,\\Bratislava, Slovakia\\\email{lukas.gajdosech@fmph.uniba.sk}}

\maketitle              
\vspace{-2em}
\begin{abstract}
    Deictic gestures, like pointing, are a fundamental form of non-verbal communication, enabling humans to direct attention to specific objects or locations. This capability is essential in Human-Robot Interaction (HRI), where robots should be able to predict human intent and anticipate appropriate responses. In this work, we propose the \textbf{M}ulti-\textbf{M}odality \textbf{I}nter-\textbf{T}rans\textbf{F}ormer (\textbf{MM-ITF}), a modular architecture to predict objects in a controlled tabletop scenario with the NICOL robot, where humans indicate targets through natural pointing gestures. Leveraging inter-modality attention, MM-ITF maps 2D pointing gestures to object locations, assigns a likelihood score to each, and identifies the most likely target. Our results demonstrate that the method can accurately predict the intended object using monocular RGB data, thus enabling intuitive and accessible human-robot collaboration. To evaluate the performance, we introduce a patch confusion matrix, providing insights into the model's predictions across candidate object locations.\\ 
    \textbf{Code} available at: \url{https://github.com/lucamuellercode/MMITF}.

\end{abstract}

\vspace{-1cm}
\section{Introduction}

The development of robots that collaborate with humans is increasing in various fields like industrial automation, healthcare, and domestic environments. As robots integrate further into society, their ability to react to human intent becomes a critical aspect of effective Human-Robot Interaction (HRI). Deictic gestures, such as pointing, provide a natural and intuitive means for humans to convey intent toward specific objects~\cite{KITA}. Due to the inherent ambiguity in natural language, pointing gestures offer a more precise spatial referral~\cite{TOMASELLO}, 
and bypass language barriers~\cite{HEWE}. According to Lenz~\cite{LENZ}, objective pointing enables humans to direct attention to objects within the shared visual field of the pointer and receiver, relying on the pointer's body pose alignment to the intended target. However, predicting a pointing target is challenging due to the need to detect hand poses, estimate direction, and identify the intended object. Traditional methods rely on measuring~\cite{HAQUE,HU,AZARI} or estimating~\cite{BAMANI} a pointing vector from 3D body representations, requiring costly hardware or extra processing, thus motivating the need for lightweight 2D approaches like ours.

\begin{figure}[t!]
\includegraphics[width=\textwidth]{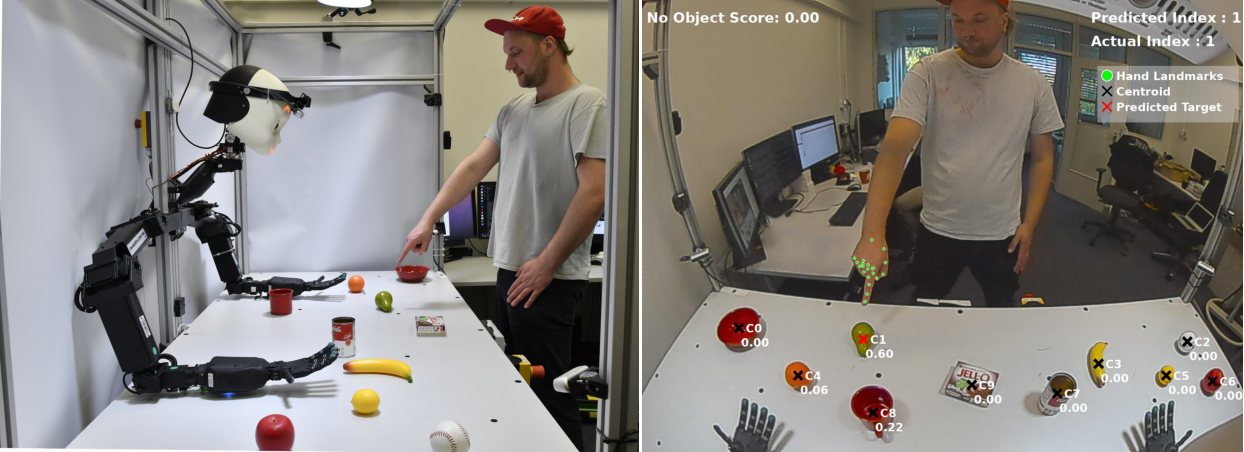}
\caption{The participant interacts with the robot, pointing to an object. Objects are represented by centroids, with scores indicating their likelihood of being the target. The probability for the non-pointing case appears in the upper left corner.}
\label{setup}
\vspace{-4ex}
\end{figure}

The field of Human-Object Interaction (HOI) offers an alternative perspective on pointing gestures, emphasizing contextual relationships in predicting human intentions when interacting with objects. As such, a pointing gesture can be considered a form of HOI in itself. Encoder-decoder transformer architectures are a useful tool to learn the contextual representations of such interactions, producing robust attention maps~\cite{ANTOUN}. Ji et al. extended this by introducing inter-modality attention~\cite{JI}, which fuses information across modalities, thus improving human-object reasoning. In our work, we propose the \textbf{M}ulti-\textbf{M}odality \textbf{I}nter-\textbf{T}rans\textbf{F}ormer (\textbf{MM-ITF}), which is a transformer-based encoder-decoder model that adapts inter-modality attention to capture the relationship between hand pose key points and object locations in a tabletop scenario (Figure~\ref{setup}). The proposed approach is RGB-based. Thus, it requires no extra equipment, wearable devices, or calibration. Our main contributions are as follows:

\vspace
{-1ex}

\begin{enumerate}
    \item An \textbf{end-to-end transformer model} (Section~\ref{sec:inter_modality_transformer}), using inter-modal attention to predict in a single forward pass whether a human is pointing, and if so, which object is targeted. 
    \item An \textbf{evaluation with a social robot} (Section~\ref{exres}) in comparison to a 2D baseline in a controlled tabletop scenario, following prior setups~\cite{ALI,ALLGEUER}.
    \item A \textbf{novel patch confusion matrix} (Section~\ref{sec:measuring_spatial_understanding}), constructed by mapping predicted object locations to discrete image regions, providing a structured visualization of the model's predictions, and supporting interpretability.
    
\end{enumerate}

\section{Related Work}
Many traditional approaches estimate a vector to determine a pointing gesture’s direction and project it into the scene to predict the target through intersection. This often requires 3D scene representations for precise estimation. Such representations can be obtained using sensors like IMUs~\cite{SIKERIDIS} and EMG~\cite{HAQUE}, offering high accuracy but requiring calibration and restricting movements. Multi-camera setups~\cite{KURAMOCHI} and depth sensors~\cite{TÖLGYESSY} can reconstruct 3D human poses, but demand additional hardware and calibration. Moreover, single RGB camera approaches infer depth using models like MiDaS~\cite{RANFTL} and pose estimators~\cite{LUGARESI,CAO,MAJI}, combining techniques as in~\cite{MEDEIROS}, where skeleton data and ORB-SLAM~\cite{MURARTAL} were integrated. However, a key challenge in pointing vector estimation is that the key points used to measure the vector—such as the forearm~\cite{HAQUE}, index finger~\cite{HU}, or a vector formed from the nose to the index finger~\cite{AZARI}—may not be collinear with the intended target. Approaches, like Bamani et al.~\cite{BAMANI}, address this by learning a pointing vector directly from 2D input, yet their approach relies on multiple models, including depth estimation, arm segmentation, and pretrained pointing estimation with wearable sensor data. Ultimately, these methods determine the target object using geometric rules, typically by computing where the pointing vector intersects a plane in the scene’s representation~\cite{MEDEIROS,BAMANI}.

On the other hand, HOI detection localizes humans and objects in a scene and predicts their interactions~\cite{ANTOUN}. Recent approaches leverage transformer-based encoder-decoder architectures~\cite{KIM,CHEN,ZOU,TAMURA,ZHANGHOI} for end-to-end scene understanding, typically following a three-step strategy. First, a backbone network, like a CNN or DETR~\cite{CARION}, extracts visual features. These feature maps are processed by the encoder for visual embedding. Then, the decoder attends to the encoded features using learnable queries representing interaction instances, followed by a simple prediction layer that generates the final HOI output: (human, interaction, object). Further, Ji et al.~\cite{JI} introduce inter-modality attention by leveraging transformer-based attention to model dependencies between any pair of tokens. This allows human pose features to attend to object features within the encoder, enhancing the embedding and improving the model’s ability to capture human-object relationships. In this work, we adopt the encoder-decoder strategy from HOI methods to interpret human pointing gestures in robotic scenarios. By integrating inter-modality attention~\cite{JI}, we model the global scene context using hand pose and object location features, mapping them as hand-object pairs. Our work enables robots to infer human non-verbal pointing cues in a modular architecture, while eliminating the need for predefined geometric rules.
\section{Methodology}

We propose MM-ITF, an approach designed to predict target objects indicated by pointing gestures, using a multimodal integration of hand poses and object locations. The architecture consists of a pretrained backbone for feature extraction, an encoder to capture global context, and a decoder that maps this context to hand-object pairs. Finally, a prediction layer assigns likelihood scores to each object. In the following, we describe the dataset, robotic platform, and architecture in detail.

\subsection{Dataset and Robot Platform}

\label{sec:dataset}
Our work revolves around the Neuro-Inspired COLlaborator (NICOL)~\cite{KERZEL}, earlier shown in Figure~\ref{setup}, which is a humanoid robotic platform designed to integrate both social interaction and physical collaboration. It features a stereo vision system, articulated arms, and an expressive face for multimodal interaction. The robot is fixed on a tabletop, creating a shared environment for collaboration between humans and the robot. The dataset consists of 30 videos, captured using the fisheye camera embedded in the robot's left eye, showing 18 participants pointing at several objects in a controlled scenario. Each video features ten standard YCB objects~\cite{YCB} on the table, facilitating the scenario's replication. Following the robot's request to point at random objects, each participant performed nine pointing tasks---seven times with a single object and twice bi-manually with two objects simultaneously.
In each frame in which pointing occurs, we locate hands and objects as 2D coordinates in the image space using pretrained models, resulting in a dataset of 572 samples with 356 pointing and 216 resting hands, each paired with a list of target objects. 


Since transformers require large amounts of training data, we apply data augmentation by modifying the 2D coordinates of hand key points and object locations. Specifically, we introduce mirroring, eight random shifts along both the x- and y-axes, and eight rotations. These transformations are applied jointly to both the hand and the objects within a sample to ensure that the hand continues to point toward the intended target. 
Additionally, we apply Gaussian noise at four increasing levels to the hand key points and object locations to introduce robustness to minor input variations. The noise is sampled from a zero-mean normal distribution with standard deviations ranging up to 3 pixels and is randomly applied to 30\% of the 2D coordinates. 
The augmented data significantly increased data variability, yielding 2,342,912 augmented samples.

\subsection{Feature Extraction and Preprocessing}
\label{FE}
Our architecture has two input channels: hand pose and object location. Given an input frame, we extract hand key points and object bounding boxes. Also, we derive a third relationship feature, representing the angle between the index finger and each object location.\footnote{We use superscripts to denote modalities: $p$ (pose), $o$ (object), and $r$ (relationship).} For hand pose estimation, we use MediaPipe~\cite{ZHANG} to detect 21 landmarks per hand. Each landmark \(\mathbf{lm}^p_i\) is a 2D coordinate:
\begin{equation} 
\mathcal{P} = \{\mathbf{lm}^p_i\}_{i=1}^{21}, \quad \mathbf{lm}^p_i \in \mathbb{R}^2
\end{equation}
The spatial arrangement of these landmarks captures what we refer to as the \textit{hand configuration}, which encodes information about the hand’s position, orientation, and gesture state, i.e., pointing or resting.

For object detection, we employ OWLv2~\cite{MINDERER}, which generates a set of bounding boxes \(\{b^o_i\}\), each defined as \((x_{\min}, y_{\min}, x_{\max}, y_{\max})\). From these, we compute their \textit{centroids} as the center point of each bounding box, forming the sequence \(\{\mathbf{c}^o_i\},\:  i \in \{1, \dots, N_t\}\), where \( N_t \) denotes the number of detected objects. To account for cases where no object is being pointed at, we define a non-object token \(\mathbf{c}_{\text{non-object}} = (-1, -1)\), choosing a value outside the valid image space for a clear distinction. With this token, the final sequence is:
\begin{equation}
\mathcal{O} = \{\mathbf{c}^o_i\}_{i=1}^{N_t} \cup \{\mathbf{c}_{\text{non-object}}\}, \quad \mathbf{c}^o_i, \mathbf{c}_{\text{non-object}} \in \mathbb{R}^2
\end{equation}

We generate a third feature as the angular alignment between the index finger landmarks and each \textit{centroid}, reflecting the relationship between each hand-object pair. The \textit{finger vector} is defined using the index fingertip and the topmost index finger joint, \(\mathbf{v}_\text{finger} = \mathbf{lm}^p_{\text{index\_finger\_tip}} - \mathbf{lm}^p_{\text{index\_finger\_dip}}\).\footnote{The index\_finger\_dip is the distal interphalangeal (DIP) joint of the index finger, the first joint below the fingertip.} For each detected object, we compute the vector from the index fingertip to the object centroid, \(\mathbf{v}_{\text{to\_centroid}, i} = \mathbf{c}^o_i - \mathbf{lm}^p_{\text{index\_finger\_tip}}\), and obtain the angle:  

\begin{equation}
\theta_i = \arccos \left( \frac{\mathbf{v}_\text{finger} \cdot \mathbf{v}_{\text{to\_centroid}, i}}{\|\mathbf{v}_\text{finger}\| \cdot \|\mathbf{v}_{\text{to\_centroid}, i}\|} \right),
\end{equation}

This results in the sequence \(\{\theta^r_i\}, \: i \in \{1, \dots, N_t\}\), where \(N_t\) denotes the number of objects. Similar to the object location sequence, we account for non-pointing hands by defining a non-relation token \(\theta_{\text{non-relation}} = -1\), chosen outside the expected valid range for radians. With this token, the final sequence is:

\begin{equation}
\mathcal{R} = \{\theta^r_i\}_{i=1}^{N_t} \cup \{\theta_{\text{non-relation}}\}, \quad \theta^r_i \in [0, \pi], \quad \theta_{\text{non-relation}} \in \mathbb{R}
\end{equation}

\subsection{Embedding and Positional Encoding}
The \( x \)- and \( y \)-values of the hand and object features are normalized to the interval \([0,1]\) using the image width \( W \) and height \( H \):

\begin{equation}
\tilde{x} = \frac{x}{W}, \quad \tilde{y} = \frac{y}{H}
\end{equation}
Since relationship, pose, and object inputs have different dimensionalities, we project them to a common embedding space of dimension \( d_T \). All \( (x, y) \) inputs, including centroid coordinates and hand landmarks, are embedded independently, with \( x \) and \( y \) projected to \( d_T / 2 \) dimensions each, and then concatenated. Angles are directly projected to \( d_T \), ensuring a unified representation across all inputs.

For positional encoding, we follow~\cite{JI,CARION} and compute sinusoidal embeddings separately for \( x \) and \( y \), concatenating them to form the final representation:
\begin{equation}
	\mathcal{PE}(\tilde{x}, \tilde{y}) = \text{concat}(PE(\tilde{x}), PE(\tilde{y})),
	\label{eq:PE}
\end{equation}
\vspace{-4ex}
\begin{equation}
	PE(*)_{2i} = \sin(* / 10000^{2i/d_T}),
	\label{eq:def_pe_sin}
\end{equation}
\vspace{-3ex}
\begin{equation}
	PE(*)_{2i+1} = \cos(* / 10000^{2i/d_T}),
	\label{eq:def_pe_cos}
\end{equation}
where \(*\) represents either \( \tilde{x} \) or \( \tilde{y} \). Since angles do not represent positional information in image space, we normalize them to \([0, 2\pi]\), project them directly to \( d_T \), and exclude them from positional encoding. After embedding and encoding, the final transformer input is structured as follows:

\begin{equation}
\mathcal{P'} = \{\mathcal{PE}(W_h \mathbf{lm}^p_i) \}_{i=1}^{21}, \quad W_h \in \mathbb{R}^{d_T \times 2}
\end{equation}
\vspace{-3ex}
\begin{equation}
\mathcal{O'} = \{\mathcal{PE}(W_o \mathbf{c}^o_i) \}_{i=1}^{N_t+1}, \quad W_o \in \mathbb{R}^{d_T \times 2}
\end{equation}
\vspace{-3ex}
\begin{equation}
\mathcal{R'} = \{W_r \mathbf{\theta}^r_i \}_{i=1}^{N_t+1}, \quad W_r \in \mathbb{R}^{d_T \times 1}
\vspace{-4ex}
\end{equation}

\begin{figure}[t!]
\includegraphics[width=\textwidth]{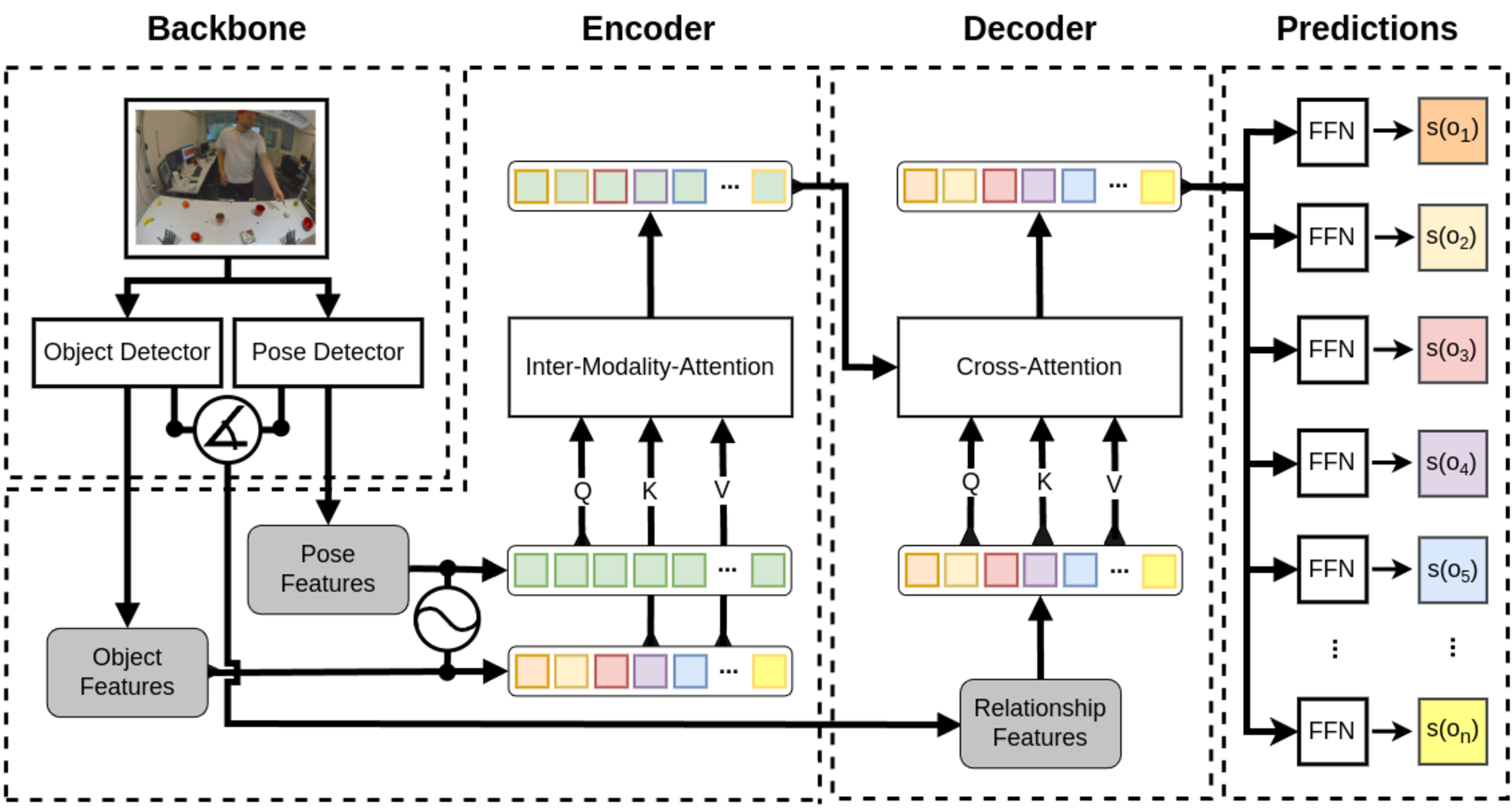}
\caption{MM-ITF combines hand pose, object locations, and their angular relationship (\protect\smallangle{45}{}) to predict pointing targets. The encoder uses hand pose features as queries (\( Q \)) and object features as keys (\( K \)) and values (\( V \)), enabling inter-modality attention to capture global context. The decoder maps this context to hand-object pairs as relationship tokens, and a Feedforward Network (FFN) assigns scores \( s(o_i) \) to all objects.}

\label{archi}
\vspace{-3ex}
\end{figure}

\subsection{Inter-Modality Transformer}
\label{sec:inter_modality_transformer}
An overview of the architecture is shown in Figure~\ref{archi}.  
Our backbone produces an object sequence \(\mathcal{O'}\) and a pose sequence \(\mathcal{P'}\) as input to the encoder. In the inter-modality attention block, pose features act as \textit{queries}, attending to object locations, which serve as \textit{keys} and \textit{values}. Each token in the pose sequence aggregates object location information, and the encoder outputs a \textit{pose-object memory} that encodes the global context between the hand and detected objects.

Our decoder processes a sequence of relationship tokens \(\mathcal{R'}\), constructed from pose and object location data. These tokens first undergo self-attention before attending, as \textit{queries}, to the \textit{pose-object memory} produced by the encoder. The cross-attention mechanism enables each relationship token to integrate scene-wide information, leveraging the global context captured by the encoder and mapping it to its respective hand-object pair. The decoder outputs a sequence of tokens, each encoding pose-object information for a specific hand-object pair.

The decoder’s output sequence is processed by a Feedforward Network (FFN) with a sigmoid activation function, assigning a score to each token. The \(i\)-th decoder output token corresponds to the \(i\)-th input relationship token, representing a specific hand-object pair. These scores allow ranking the objects, where the model predicts the index \( j \) of the token with the highest score. Since each token retains its mapping to the original input relationship tokens, the predicted index identifies the hand-object pair most likely to fulfill the pointing relationship.

We frame our task as binary classification, where each output representation is evaluated based on whether it fulfills the pointing relationship. To optimize this, we use Binary Cross-Entropy (BCE) loss, encouraging the model to assign higher scores to hand-object pairs that align with the pointing relationship while reducing scores for those that do not. 
Since our goal is to generate scores for every object rather than make a strict binary decision, we do not apply a threshold to separate pointing and non-pointing tokens.  Instead, we rank the raw scores, selecting the object with the highest likelihood as the predicted pointing target.

\vspace{-2ex}
\section{Experiments and Evaluation}
\vspace{-1ex}
We evaluate the MM-ITF architecture by comparing different channel configurations. Specifically, we compare a two-modality setup utilizing the hand pose and object location and a three-modality setup consisting of the same modalities in addition to the relationship feature (see~Section~\ref{FE}). Our two-modality setup uses the same architecture described in Figure~\ref{archi} where the relationship feature is replaced by the object location as input to the decoder. The results of both modality configurations are compared to a baseline that predicts objects based on proximity to a vector derived from the pointing gesture. In the following, we introduce the baseline, present the results, and conclude with a visual analysis of the model’s predictions using a patch confusion matrix. 

\subsection{Baseline for Evaluation}
As baseline, we use a 2D method proposed by Ali et al.~\cite{ALI} to predict objects indicated by pointing hand gestures. This method was applied to the same tabletop scenario with the NICOL robot. It consists of a Multi-layer Perceptron (MLP) that uses hand landmarks to determine whether the user is pointing. The final target objects are predicted based on proximity of the object's centroid to a continuous line that passes through the wrist and index finger key points. We choose this baseline because, similar to our approach, it relies completely on 2D data. Furthermore, its prior application within the same robotic setup ensures a fair and meaningful comparison. The baseline uses the same pretrained models for extracting hand key points and object locations, i.e., MediaPipe and OWLv2.

\subsection{Experiment Results}
\label{exres}
\begin{table}[t!]
	\centering
    \caption{Average object prediction results for the baseline method and our proposed MM-ITF architecture, with a two-modality and a three-modality configuration}
	\begin{tabularx}{\columnwidth}{lYYYYc}
		\toprule
		\textbf{Model} & \textbf{Accuracy} & \textbf{Precision} & \textbf{Recall} & \textbf{F1-Score} & \textbf{Top-2 Accuracy} \\
		\midrule
		Baseline~\cite{ALI} & 0.89 $\pm 0.008$ & 0.84 $\pm 0.007$ & 0.90 $\pm 0.012$ & 0.85 $\pm 0.008$& \textbf{0.96} $\pm 0.004$ \\
		MM-ITF* & 0.71 $\pm 0.044$ & 0.70 $\pm 0.037$ & $0.68 \pm 0.044$ & 0.67 $\pm 0.041$& 0.92 $\pm 0.014$\\
		MM-ITF** & \textbf{0.90} $\pm 0.017$ & \textbf{0.88} $\pm 0.019$ & \textbf{0.92} $\pm 0.019$& \textbf{0.90} $\pm 0.019$& \textbf{0.96} $\pm 0.008$ \\

		\bottomrule
    \multicolumn{6}{l}{\scriptsize  * \hspace{0.1em} Our model trained with \textbf{two modalities} (hand pose and object locations).} \\
    \multicolumn{6}{l}{\scriptsize ** Our model trained with all \textbf{three modalities}, including the relationship feature.} \\
	\end{tabularx}
	\label{tab:averages_performance}
\vspace{-3ex}
\end{table}

\vspace{-1ex}

We performed an eight-fold cross-validation, in which we trained eight models for both the baseline gesture classifier and our MM-ITF architecture. The 30 scenes from the dataset (see Section~\ref{sec:dataset}) were split for training, validation, and testing. Each model was trained on 21 scenes and validated on a unique subset of three, ensuring that all scenes were used for validation once, except for six scenes which were held out as a test set. Table~\ref{tab:averages_performance} summarizes the test results, in which the values reflect the average performance across all models. MM-ITF in the three-modality setup (pose, object, and relationship data) achieves 90\% accuracy, slightly above the baseline at 89\%, showing that objects indicated by pointing gestures can be learned to a comparable performance without additional geometric post-processing. While the baseline relies on a two-stage geometric approach, our model jointly predicts both the gesture state and target object in a single step. Both methods similarly achieve high Top-2 accuracy, ranking the target object within the top two choices in 96\% of cases.  

Although the MM-ITF two-modality setup (pose and object) reaches only 71\% accuracy, it achieves a high Top-2 accuracy of 92\%, showing that it learns a meaningful link between the hand pose and object location but struggles to make a fine-grained correct final prediction. This suggests that while the model captures contextual relationships between hand pose and object location, it benefits noticeably from the relationship feature to improve object ranking precision.

Since the baseline selects the nearest object along a continuous line through the wrist and index finger, its predictions are straightforward to interpret. However, we need a confusion matrix to analyze the MM-ITF's performance. A confusion matrix over predicted indices would offer limited insight, as object locations vary across samples, and constructing one based on centroids is impractical due to the extremely large number of possible positions. Therefore, we introduce a method to discretize our architecture's outputs, mapping object centroids to fixed image regions for structured visual analysis.

\begin{figure}[t!]
\includegraphics[width=\textwidth]{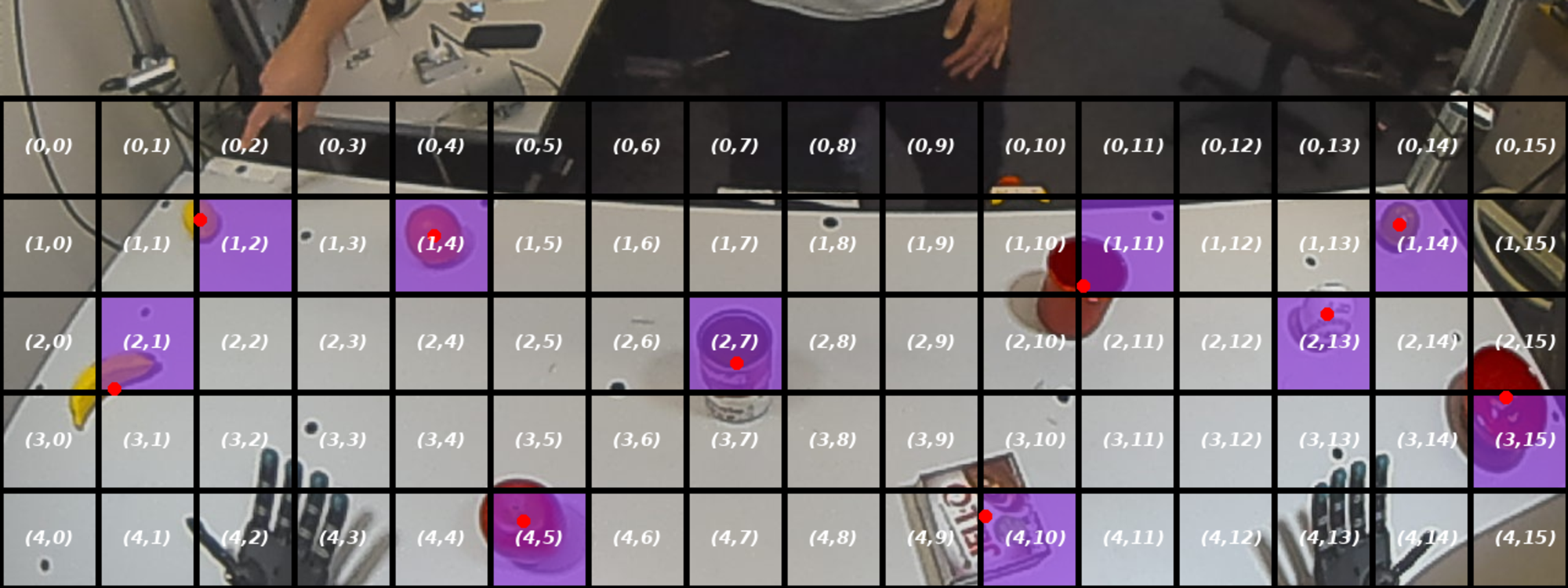}
\caption{A visualization of the performance of our architecture using patches. The table space is divided into evenly sized, non-overlapping patches, and centroids (red dots) are assigned to patches by dividing their \(x, y\) coordinates by the patch width and height. The assigned patches are highlighted in purple.} \label{ptc}
\vspace{-3ex}
\end{figure}

\subsection{Measuring Spatial Understanding}
\label{sec:measuring_spatial_understanding}


We further evaluate the performance of our architecture by visualizing the predicted objects. The table area in the image space is divided into evenly sized, non-overlapping patches, as illustrated in Figure~\ref{ptc}. Since our model predicts an index corresponding to the object centroid \((x, y)\) in image coordinates, each predicted and ground-truth centroid is mapped to a patch based on its coordinates. This mapping is achieved by dividing the \(x\)-coordinate by the patch width and the \(y\)-coordinate by the patch height. As a result, the centroid predictions are discretized into predefined image regions, enabling structured spatial analysis through a confusion matrix over patches. Patches with no assigned predictions are filtered out, and row normalization is applied to enhance interpretability.

For a more detailed analysis of our MM-ITF model's output, we construct a patch confusion matrix, which visualizes how predicted object centroid locations align with ground-truth centroids. Assigned patches for predicted centroids are shown along the x-axis, and those for ground-truth centroids along the y-axis. Each entry \((i, j)\) represents how often a ground-truth centroid in patch \(i\) is predicted as patch \(j\). Diagonal entries correspond to correct predictions, while off-diagonal values indicate spatial misclassifications. Additionally, the first row and last column of the matrix represent the non-object class, distinguishing non-pointing gestures from those associated with an object.



\begin{figure}[h]
\centering
\includegraphics[width=0.8\textwidth]{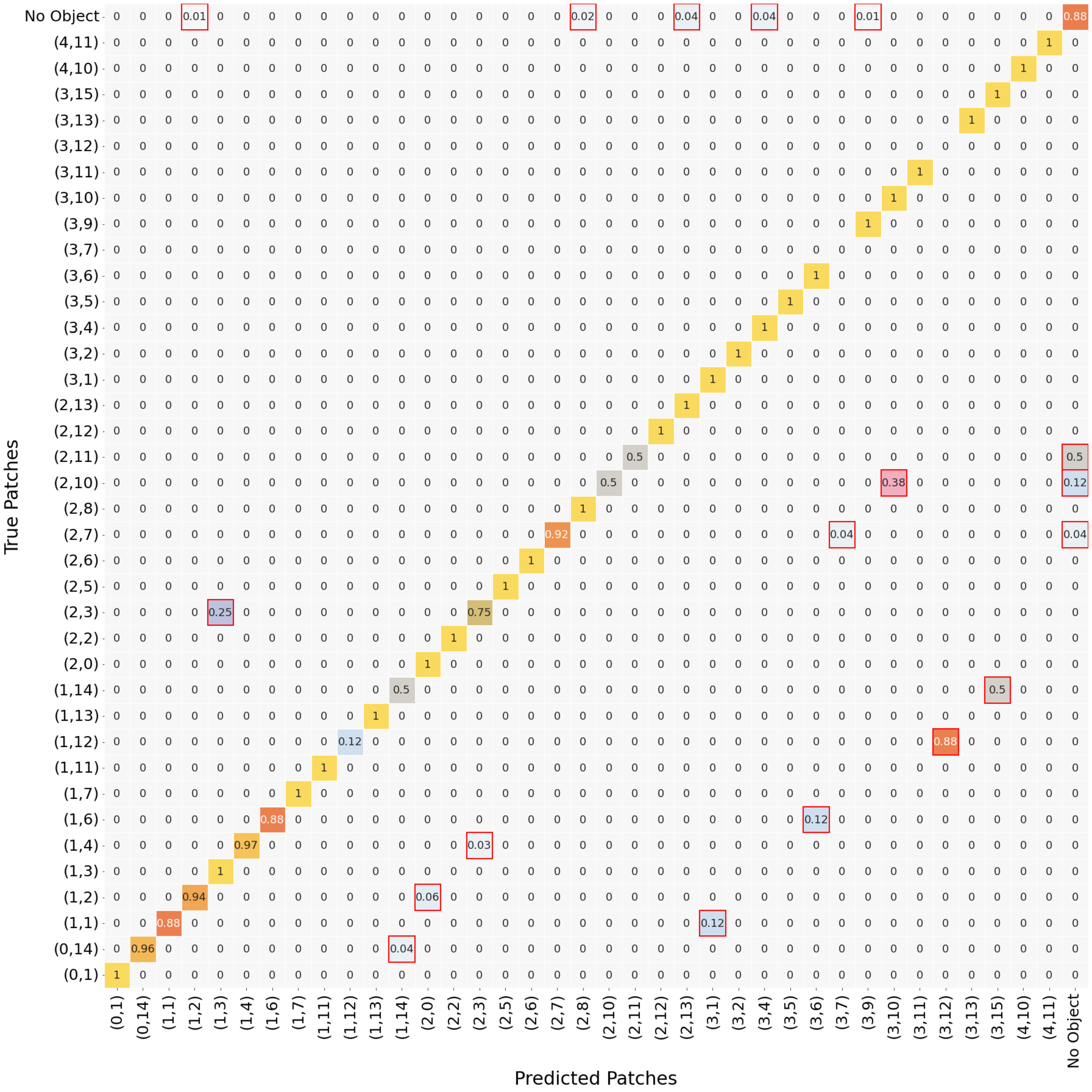}
\caption{The patch confusion matrix shows the mapping between predicted and target centroids to discrete regions in image space. Predictions are shown on the x-axis, targets on the y-axis. The first row and last column represent the non-object case.}
\label{PCM}
\end{figure}

Examining the confusion matrix (Figure~\ref{PCM}) for the three-modality setup reveals that correct predictions frequently receive high scores, demonstrating the model’s ability to distinguish target objects of a pointing gesture, even in close proximity.\footnote{The matrix shows average results over eight models; see Section~\ref{exres} for details.} This suggests that the relationship feature plays a key role in guiding target selection. However, when multiple objects align perfectly with the pointing direction, the model often predicts an object behind the actual target relative to the participant. For example, objects at (1, 12) are misclassified as (3, 12) in 88\% of cases, (2, 10) as (3, 10) in 38\%, and (1, 14) as (3, 15) in 50\% of cases. Together, these findings indicate a strong reliance on hand-object alignment.

Beyond object predictions, the model distinguishes between pointing and non-pointing hands but prioritizes hand location over \textit{hand configuration}. For example, it sometimes predicts no-object for gestures originating from positions typically associated with resting hands, as observed at patch (2, 11). This suggests that hand position outweighs landmark arrangement, reinforcing the model’s reliance on hand-object alignment and proximity over hand articulation.
The results for the two-modality setup reinforce our findings. The relationship feature improves both target object and gesture state identification, reducing confusion between closely positioned objects and correctly classifying aligned ones. It also helps distinguish pointing gestures from resting hand positions, reducing misclassification as non-pointing. While the two-modality setup reaches a high Top-2 accuracy of 92\%, indicating that the model captures general spatial relations, the relationship feature provides crucial guidance for fine-grained distinctions and prevents over-reliance on proximity cues.

\vspace{-2ex}

\section{Conclusion}
In this work, we proposed a framework for interpreting human pointing gestures toward objects. Operating purely in 2D, we leveraged the transformer's attention mechanism as a coherence score to map deictic gestures to object locations—without relying on predefined geometric rules, additional equipment, or 3D representations of the shared workspace. Our approach predicts, in a single step, whether a user is pointing and, if so, the most likely target object. This enables a reliable mapping between a deictic gesture and its inferred object, based solely on human pose. By ranking likely target objects, our method serves as a building block for downstream tasks aimed at estimating human intent in collaborative scenarios. In this sense, our work contributes to a robot’s social skill set, enabling more intuitive and seamless interaction through the interpretation of pointing gestures. The architecture is modular and extendable, reflecting the multimodal nature of human communication.

While the current setup assumes fixed positions for the camera, table, and participants, future work will explore more dynamic and flexible interaction settings. We also aim to extend the hand features with modalities such as gaze, further enriching the global context between human and object features modeled by the encoder—and, in doing so, expanding the possibilities for humans to effortlessly indicate their intent to a robot.

\begin{credits}

\subsubsection{\discintname}
The authors declare that they have no conflict of interest.

\end{credits}
%
%
%

\bibliographystyle{splncs04}
\bibliography{references}

\end{document}